\def\year{2020}\relax
\documentclass[letterpaper]{article} 
\usepackage{aaai20}  
\usepackage{times}  
\usepackage{helvet} 
\usepackage{courier}  
\usepackage[hyphens]{url}  
\usepackage{graphicx} 
\usepackage{graphicx}  
\usepackage[normalem]{ulem}

\usepackage{xcolor} 

\usepackage{subcaption}

\makeatletter
\newcommand{\thickhline}{%
    \noalign {\ifnum 0=`}\fi \hrule height 1pt
    \futurelet \reserved@a \@xhline
}

\title{Context-Based Quotation Recommendation\thanks{Paper will appear in the proceedings of 15th International AAAI Conference on Web and Social Media (ICWSM 2021).}}
\author{
Ansel MacLaughlin$^{1}$\thanks{Work completed while interning at Google Research.}\qquad 
Tao Chen$^{2}$\thanks{Corresponding author.}\qquad 
Burcu Karagol Ayan$^{2}$\qquad
Dan Roth$^{3}$\\
$^{1}$ Northeastern University, Boston, MA, USA\\
$^{2}$ Google Research, Mountain View, CA, USA \\
$^{3}$ University of Pennsylvania, Philadelphia, PA, USA \\
ansel@ccs.neu.edu \quad \{taochen, burcuka\}@google.com \quad danroth@cis.upenn.edu
}

\begin{document}

\maketitle

\begin{abstract}

While composing a new document, anything from a news article to an email or essay, authors often utilize direct quotes from a variety of sources. Although an author may know what point they would like to make, selecting an appropriate quote for the specific context may be time-consuming and difficult. We therefore propose a novel context-aware quote recommendation system which utilizes the content an author has already written to generate a ranked list of quotable paragraphs and spans of tokens from a given source document. 

We approach quote recommendation as a variant of open-domain question answering and adapt the state-of-the-art BERT-based methods from open-QA to our task. We conduct experiments on a collection of speech transcripts and associated news articles, evaluating models' paragraph ranking and span prediction performances. Our experiments confirm the strong performance of BERT-based methods on this task, which outperform bag-of-words and neural ranking baselines by more than $30\%$ relative across all ranking metrics. Qualitative analyses show the difficulty of the paragraph and span recommendation tasks and confirm the quotability of the best BERT model's predictions, even if they are not the true selected quotes from the original news articles.

\end{abstract}

\section{Introduction}
\label{sec:introduction}

Machine-assisted writing tools have many potential uses, including helping users compose documents more quickly \cite{Hard2019}, generate creative story ideas \cite{Clark2018b}, and improve their grammar \cite{RozovskayaRo14,Hoover2015}. However, writing a document, such as a news article, academic essay, or email, consists not only of composing new text, but of weaving in content from source documents by summarizing, paraphrasing, and quoting. Quotes are a particularly important part of many documents, functioning to ``change the pace of a story, add colour and character, illustrate bald facts, and introduce personal experience'' \cite{Cole2008}. However, the source documents from which an author could select a quote may be long, consisting of hundreds of paragraphs, and, as described in the findings of our pilot study (\S\ref{sec:pilot}), searching through them to select an appropriate quote may be time-consuming and challenging.

\begin{table}[t]
	\caption{Given a title and context in a new document, models aim to recommend both relevant source paragraphs and quotable spans of text in them, such as in this example. The \textbf{bold} portion of paragraph was quoted by the author in the next sentence of the document.}
	\small
	\begin{tabular}{|p{8cm}|}
		\hline
		\textbf{Document Title:} Return engagement to White House for Miami Heat        \\ 
		\textbf{Document Context:} ... The president opened by recounting the numerous accomplishments of the team last season, including Miami's team-record 66 wins.
		$\mathtt{[Recommend}$ $\mathtt{Quote}$ $\mathtt{Here]}$ 
		\\ 
		\textbf{Source Paragraph:} Last season, the Heat put together one of the most dominating regular seasons ever by a defending champion. They won a team-record 66 games. At one point, they won 27 games straight- the second-longest winning streak ever, \textbf{extraordinarily impressive- almost as impressive as the Bulls' 72-win season}. Riley and I were reminiscing about those Knicks years.	
		\\ \hline
	\end{tabular}
	\label{table:example_datapoint}
\end{table}

In this paper we propose a novel context-aware quote recommendation (QR) task -- recommending quotable passages of text from a source document based on the content an author has already written. Previously proposed QR systems have either made predictions without using the textual context \cite{Niculae2015} or only predicted from a fixed list of popular quotes, proverbs, and maxims \cite{Lee2016,Tan2015,Tan2016,Ahn2016,Tan2018b}. Our proposed task is more general and difficult, as models must both identify quotable passages of arbitrary length in any given source document and recommend the most relevant ones to the author. Authors across a variety of domains could benefit from such a system, including journalists quoting from an interview, speech or meeting transcript, students quoting from a chapter of a primary source for an essay, or social media posters writing about a recent blog post or news article.

We approach the problem at two different granularities, recommending both entire paragraphs in a source document for an author to excerpt and specific spans of text in those paragraphs that are suitable for quoting (i.e. quotable). Table \ref{table:example_datapoint} shows an example recommendation -- given the title and a snippet of context from a new document, we aim to recommend both a relevant and quotable source paragraph and specific subspan that an author can quote in the next sentence of their document. The two granularities of recommendation offer different and complementary functionality to authors. Recommending relevant and quotable source paragraphs would allow authors to find suitable quotes more quickly, potentially searching over only the top ranked paragraphs for a span of tokens to excerpt. Further, since, as discussed in the findings of our pilot study (\S\ref{sec:pilot}) and error analysis (\S\ref{sec:results}), authors often paraphrase surrounding content from the paragraph in the introduction to and discussion of a quote, the paragraph's entire text beyond the specific span is still useful. Recommending spans of text, although more difficult than paragraphs, would save an author further time by allowing them to easily scan through a ranked list of potential quotes.

We cast our novel approach to QR as a variant of open-domain question answering (open-QA). In open-QA, for a given question, a model must retrieve relevant passages from a source corpus and identify answer spans in those passages \cite{Chen2017}. Similarly, for our task, for a given context in a new document, a model must both identify relevant paragraphs in a source document and predict quotable spans of tokens in them. Following this intuition, we adapt state-of-the-art (SoTA) BERT-based open-QA methods for our task \cite{Devlin2019}. We fine-tune two separate BERT models, a paragraph ranker and a span predictor \cite{Wang2019}, and combine their predicted scores to rank source paragraphs and spans by their predicted quotability.  

We conduct experiments on a dataset of White House speech and press conference transcripts and associated news articles \cite{Niculae2015}. To first assess the difficulty of the QR task, we recruit two journalists for a pilot study. We find that identifying the ``correct'' (actually quoted) speech paragraphs and spans is hard, even for experts, and that, unlike QA, there may be multiple relevant and quotable paragraphs and spans for a given context. Next, we adapt and apply SoTA BERT-based open-QA methods to the task. We find that the combined BERT model \cite{Wang2019} performs best, significantly outperforming bag-of-words and neural ranking baselines. Finally, in light of the findings of our pilot study, we perform a large-scale crowdsourced evaulation of the quality of our best model's top recommendations, focusing on instances where the model misranks the paragraph chosen by the original author. We find that human raters judge the model's top predicted (non ground truth) paragraph to be relevant and quotable 85\% of the time, further demonstrating the usefulness of our system.

Our contributions include a novel approach to QR that generalizes to source documents not seen during training, an application of SoTA BERT-based open-QA methods that combine paragraph and span-level models to capture complementary aspects of a passage's quotability, and a qualitative evaluation methodology aimed at better dealing with the fact that quote selection is subjective.

\section{Related Work}
\label{sec:related_work}

\subsubsection{Quotability Identification \& Quote Recommendation} 
Our proposed task combines quotability identification, identifying and extracting quotable phrases from a source document, and quote recommendation, recommending quotes to an author as they compose a new document. No earlier work has addressed both of these problems simultaneously. Prior quotability identification research has focused only on identifying generally quotable content, regardless of the context in which its used \cite{Tan2018,Bendersky2012,Mizil2012,Niculae2015}. On the other hand, previously proposed QR models do not identify quotable content -- they only make recommendations from a predefined list of quotable phrases \cite{Tan2015,Tan2016,Lee2016,Ahn2016,Tan2018b}.

Unlike the proposed BERT models, previous quotability identification systems have used linear classifiers with manually designed lexical and syntactic features to distinguish between quotable and un-quotable passages. Prior work has studied applications in a variety of domains, including political speeches and debates \cite{Tan2018,Niculae2015}, books \cite{Bendersky2012} and movie scripts \cite{Mizil2012}.

Both collaborative filtering and content-based approaches have been used for QR systems. Unlike our work, which recommends quotes to a specific context in a new document, \citeauthor{Niculae2015} \shortcite{Niculae2015} approach the problem as matrix completion and learn to predict which quotes a given news outlet will report. They do not use textual features for prediction and thus cannot generalize to new, previously unseen, quotes. 

More similar to our work are content-based QR systems that use textual content to recommend quotes for a given context in a new document \cite{Tan2015,Tan2016,Lee2016,Ahn2016,Tan2018b}. However, prior work approach the problem not as ranking paragraphs and predicting spans in a source document, but as ranking a predefined list of popular quotes, proverbs and maxims. Furthermore, unlike BERT, previous content-based systems neither deeply model the interactions between a context and quote nor identify the textual features in a quote indicative of its overall quotability.
Finally, there are substantial modeling differences between our proposed system, which uses only the textual content of both the context and paragraph to make a recommendation, and prior work. \citeauthor{Tan2015} \shortcite{Tan2015,Tan2016,Tan2018b} train their models with various additional metadata features, such as topic and popularity data crawled from the web. 
\citeauthor{Lee2016} \shortcite{Lee2016} and \citeauthor{Ahn2016} \shortcite{Ahn2016} approach recommendation as multi-class classification, using only the text of the context to predict a specific quote class. These differences make previous systems incompatible with our proposed task, where metadata may not be available and models must rank an arbitrary number of source paragraphs not seen during training.

\subsubsection{Citation Recommendation}

Our proposed QR system has similarities to content-based citation recommendation (CR) systems. These systems recommend citations for a new document based on its textual similarity to the candidate citations. Global recommendation systems \cite{Bhagavatula2018,Strohman2007,He2010,Gupta2017}
use the entire text of the new document to recommend candidate documents to cite. Our proposed task is most similar to local CR, where models condition their recommendations on only a small window of tokens near the value to be predicted rather than the entire new document \cite{Yu2011,Huang2015,He2010,He2011,Peng2016,Fetahu2016,Jeong2019}. However, unlike CR, QR models must learn to make finer-grained recommendations, identifying both paragraphs and arbitrary length subspans in a source document that might be most relevant and quotable. Thus, although existing CR systems could be applied to the paragraph ranking portion of our QR task by treating each source paragraph as a different document, they would not be able to also identify and recommend specific quotable spans of tokens in those paragraphs.  

Most prior CR work has been applied to the domain of scientific articles, recommending citations for an academic paper, but applications to news data, predicting which news articles will be cited in a given context, have also been explored \cite{Fetahu2016,Peng2016}. Recently, neural networks have shown strong performance on citation and other content-based recommendation tasks, outperforming traditional bag-of-words and citation translation models \cite{Huang2015,Bhagavatula2018,Ebesu2017,Zhang2018,Jeong2019}. Currently, BERT-based models achieve SoTA performance on local recommendation problems \cite{Jeong2019}. Though the BERT-based model of \citeauthor{Jeong2019} \shortcite{Jeong2019} uses additional citation network features to make predictions, making it incompatible with our context-only QR task and unable to generalize to new source documents not seen during training, the strong performance of BERT further motivates our model selection for QR.

\subsubsection{Question Answering}
\label{sec:related_qa}

As described earlier, we cast our proposed QR task as a variant of open-QA. For a given question, open-QA systems first retrieve $d$ relevant documents, usually with traditional IR models such TF-IDF \cite{Chen2017,Lee2018,Clark2018a} or BM25 \cite{Wang2018a,Wang2018b,Wang2019}. To decrease the search space for the machine reader, many systems train a paragraph or sentence-level ranker to estimate the likelihood that a passage contains the answer \cite{Clark2018a,Htut2018,Lee2018,Wang2018a,Wang2018b,Wang2019}. Finally, a machine reader is trained to score answer spans in the top $p$ ranked paragraphs. At prediction time, the reader processes and predicts a span in each of the $p$ paragraphs independently, and the overall confidence of each predicted span is calculated as a combination of the scores from the document ranker, paragraph ranker and reader. Recently, BERT \cite{Devlin2019} has shown strong performance on both open-QA \cite{Liu2019,Wang2019} and single paragraph QA datasets, such as SQuAD \cite{Rajpurkar2016}, 
where no document or paragraph ranking is required \cite{Alberti2019}. For our QR task, we adapt SoTA BERT-based methods from open-QA \cite{Wang2019}, combining the scores of two separately trained BERT models, a paragraph ranker and span predictor, to rank paragraphs and spans.

Though the methodological solutions for QA and our proposed QR task are identical, there are differences in how end users would use model predictions. In QA, for a given question and set of documents, there are known paragraph(s) in the documents which contain the correct answer(s). Ranking the paragraphs is only an intermediate step that assists in finding the correct answer span. End users are expected to only read the predicted answer, not the entire paragraph. However, as described in the findings of our pilot study (\S\ref{sec:pilot}) and crowdsource evaluation (\S\ref{sec:crowdsource}), in QR there are often multiple paragraphs and spans an author could reasonably choose in a given context. Retrieving topically relevant and quotable paragraphs is therefore an equally important task, as authors may read multiple top ranked paragraphs and  their suggested spans as they search for a quote.

\section{Quote Recommendation}
\label{sec:quote_rec}

At a high level, QR is a content-based recommendation task -- recommending relevant and quotable content from a source document to an author as they compose a new document. Specifically, given a snippet of context in a new document and the text of a \textit{single} source document that the author wants to quote, we learn both to rank the paragraphs of the source document by their relevance and quotability and to extract quotable spans of text from those paragraphs. To give the most relevant recommendations, this snippet of context should contain the content an author has written up to the point they are recommended and select a quote. Thus, we include in each context the sentences in the new document that \textit{precede} the quote. We henceforth refer to this as the \textbf{left context}. We do not include the part of the sentence containing the quote or content after, since those often contain analysis of the quote's content. Table \ref{table:example_datapoint} shows a portion of an example left context. The bold portion of the source paragraph is then quoted in the \textbf{next sentence} of the new document. We also include the new document's title as further context as it is acts as compact summary and allows us to make context-based predictions when there is no left context (i.e. the quote occurs in the first sentence). We refer to the concatenation of the new document's title with the quote's left context in the document as the \textbf{quote query}. 

As described in \S\ref{sec:related_work}, our proposed QR task differs from previously proposed QR task variants \cite{Niculae2015,Tan2015,Tan2016,Lee2016,Ahn2016,Tan2018b}. For our task, systems must both rank the candidate source paragraphs and learn to identify specific quotable spans of text in them. Prior context-based QR task variants do not include this span prediction component. Instead, they cast QR as a simply ranking a predefined list of quotes, proverbs and maxims \cite{Tan2015,Tan2016,Lee2016,Ahn2016,Tan2018b}.

\subsection{Pilot Study: Expert Performance}
\label{sec:pilot}
To assess the difficulty of our proposed QR task, we conduct a small pilot study. We recruit two experienced journalists and ask them to perform the following two tasks:

\begin{enumerate}
	\item Paragraph selection: Given a context (title \& entire left context) in a new document and all the paragraphs in the source document, identify all of the paragraphs that are relevant and might be appropriate to quote in the next sentence of the article.
	
	\item Span selection: Given a context (title \& entire left context) in a new document and the actual paragraph in the source document that the author quoted in the next sentence, predict which text span (can be a sentence, incomplete sentence, a span of multiple sentences, etc.) was actually quoted.
	
\end{enumerate}
We are interested in examining how difficult it is for experts to find and suggest relevant and quotable source passages, and if the expert suggestions line up with the actual passages quoted in the original articles.

For our study, we sample from the test split of our dataset. As detailed in \S\ref{sec:dataset}, the source documents in this dataset are Presidential Speeches and the new documents are news articles which report on and quote from the speeches. We sample 1 speech, ``Remarks by the President at the Healthy Kids and Safe Sports Concussion Summit,'' and sample 1 quote from each of the 15 associated news articles for inclusion in our analysis. 
The journalists perform task 1 first, marking all quotable paragraphs for each of the 15 contexts. Then, using the same contexts, they perform task 2, this time also viewing the positive paragraph quoted by the original author. Finally, the journalists were asked to assess each task's difficulty and list the factors that influenced their predictions, focusing on what factors made specific paragraphs and spans relevant and quotable. 

The selected speech contains 28 paragraphs, of which 7 are quoted across the 15 articles. 3 of the 15 quotes are the same excerpt of text -- the other 12 are unique, though some overlap. For each quote in task 1, we calculate paragraph accuracy as 1 if the journalist identified the positive paragraph in their list of relevant paragraphs, otherwise 0. Since the setup for our task 2 is identical to QA, we use 2 common QA metrics: exact match (EM), and macro-averaged bag-of-words F1. EM measures if the predicted span exactly matches the positive quote, and BOW-F1 measures their average word overlap. 

\begin{table}
	\caption{Expert Analysis Results: predictive accuracy of the 2 journalists on a sample of 15 quotes from our test set.}
	\label{table:expert_analysis}
	\small
	\centering
	\begin{tabular}{cccc}
		
		&  \multicolumn{1}{c}{Paragraph} & \multicolumn{2}{c}{Span} \\
		\multicolumn{1}{c|}{Journalist} & \multicolumn{1}{c|}{Accuracy} & EM & BOW-F1 \\ \hline
		\multicolumn{1}{c|}{A}          & \multicolumn{1}{c|}{66.6}                   &      20.0              & 56.2 \\
		\multicolumn{1}{c|}{B}            & \multicolumn{1}{c|}{46.7}                    &       00.0               & 46.9
	\end{tabular}
\end{table}

As seen in Table~\ref{table:expert_analysis}, identifying the originally quoted paragraphs and spans is difficult, even for experts. On task 1, journalist A identified the positive paragraph in 10 of 15 instances and journalist B in 7. A and B listed a median of 2 (mean 3.8) relevant and quotable paragraphs per news article context. On task 2, both journalists struggled to identify the exact quote used by the original author -- A identified 3 of the exact quotes and B none. The most common error was over-prediction (74\% of errors), where the journalist predicted quotes that contained both the true quote and some extra tokens on either side (see example in Table~\ref{table:example_journalist_prediction}). This was due in part to the journalists' tendencies to predict complete sentences as quotes (70\% of predictions). However, examining the original news articles, we find that the original authors often paraphrase content from the surrounding paragraph when introducing a direct quote. Thus, over-prediction is a reasonable error since authors may simply paraphrase portions of the predicted quote.  

For feedback, the journalists noted that they only closely considered paragraphs and spans with specific anecdotes and figures that built on the previous context. Both noted that, while neither task was difficult, task 1 was harder since they had to search the entire speech and read multiple paragraphs. However, one journalist noted that task 1's search to identify relevant paragraphs became easier and faster as they went along because they started to memorize the speech. 

These findings indicate the usefulness of our proposed QR system as a writing assistance tool -- ranking paragraphs by their predicted quotability would help authors find quotes more quickly. This would help prevent authors from having to nearly memorize each source document to ensure fast recall, which might be difficult when authors are also focused on composing the new document or if the source document is long (the median speech in our dataset is 43 paragraphs, 1.5x the length of the Concussion Summit speech).  

Further, the difficulty for experts to identify the paragraphs and spans quoted by the original authors, along with the median of 2 quotable paragraphs listed per news article context, point towards the subjectivity of QR and its differences from QA. In traditional QA datasets, there is only a single correct answer for a given question. On the other hand, in QR, multiple paragraphs and quotes could be suitable for a given context. Traditional recommendation and QA metrics, therefore, may be especially harsh for evalauting the quality of a model's predictions. Motivated by this finding, in \S\ref{sec:crowdsource} we perform a crowsourced evaluation of our best model's predictions to assess whether its ``incorrect" recommendations (i.e. of passages not selected by the original authors) are still relevant and appropriate. 

\begin{table}[t]
	\caption{Example of over-prediction from our pilot study, where the predicted quote contains both the \textbf{true quote} and some extra phrases. Over-prediction is a reasonable error since authors often paraphrase portions of the surrounding content in the introduction to and discussion of a quote.}
	\small
	\begin{tabular}{|p{8cm}|}
		\hline
		\textbf{Document Title:} White House concussion summit opens with introduction from Huntingtown's Tori Bellucci \\
		\textbf{Document Context:} ... In his brief remarks Thursday, Obama discussed the importance of sports in American society, and the need to ensure the protection of youth who play them. ``We want our kids playing sports,'' Obama said.
		\\ 
		\textbf{Predicted Quote:} I'd be much more troubled if young people were shying away from sports. \textbf{As parents, though, we want to keep them safe}, and that means we have to have better information.	
		\\ \hline
	\end{tabular}
	\label{table:example_journalist_prediction}
\end{table}

\section{Methods}
\label{sec:methods}

As discussed in \S\ref{sec:introduction}, our QR problem can be viewed as a variant of open-QA, where the questions are the quote queries, the passages containing the answers are the different paragraphs of the source document, and the answers are the exact quotes the author used in the next sentence of the document. Thus, to model the QR problem, we apply SoTA BERT-based methods from open-QA. As described in \S\ref{sec:related_work}, SoTA open-QA methods consist of three parts -- 1) a bag-of-words retrieval system, such as BM25, to retrieve an initial set of candidate passages 2) a passage reranker for selecting high quality passages 3) a machine reading module which predicts an answer span in a given passage. The final score for a passage and its predicted span is computed as some combination of the scores from the three steps. Since, for our task, we are only focusing on a single source document, the initial retrieval step is skipped and we only model the paragraph ranking and span prediction tasks.

Following \citeauthor{Wang2019} \shortcite{Wang2019}, we select BERT as both our passage ranker and span predictor. In addition to achieving SoTA performance on open and single-paragraph QA \cite{Devlin2019,Liu2019,Yang2019,Wang2019}, BERT has also achieved SoTA performance on passage and document retrieval \cite{Dai2019,Yilmaz2019,Nogueira2019}. Since, unlike QA, users of a QR system are likely to examine and use the full text of multiple top ranked paragraphs, paragraph retrieval performance is more important than just as a means of achieving better span prediction.

\subsubsection{BERT Model Inputs}

The paragraph and span-level BERT models receive the same input and are identically structured except for the final layers, loss functions and example labels (paragraph or span-level labels). Due to memory constraints, BERT cannot operate on an entire source document as a single input, so we operate on the paragraph level, with BERT processing each (quote query, source paragraph) pair as a single packed input.

For input to BERT we tokenize titles, left contexts and source document paragraphs into WordPieces \cite{Wu2016} and cap them at predetermined lengths chosen as hyperparameters. Left contexts are capped from the right, so they always include the words directly preceding the quoting sentence. We treat the quote query as sequence A and candidate source paragraph as sequence B. We add a special new token `$\mathtt{[body\_start]}$' to separate the title from the left context to give the model a notion of which part of the new document it is reading \cite{Alberti2019}. Thus the WordPiece input to BERT is:\\
\centerline{$\mathtt{[CLS]}$ $\mathtt{Title}$ $\mathtt{[body\_start]}$ $\mathtt{Context}$ $\mathtt{[SEP]}$ $\mathtt{Paragraph}$ $\mathtt{[SEP]}$}\\
By packing two sequences into the same input, BERT can use self attention to model interactions between the sequences across multiple levels (word, sentence, paragraph). Since there may be many topically related paragraphs in a source document with differing amounts of quotable content, a model must also learn to judge how interesting and meaningful specific content in a paragraph is.

\subsubsection{Paragraph Ranking}


To generate training and testing examples for the models, we iterate over each quote query and create a (quote query, source paragraph) example pair for each paragraph in the corresponding source document. Each pair has a binary label: 1 if the author actually quoted from the paragraph in the next sentence of the document, 0 otherwise. Following \citeauthor{Wang2019} \shortcite{Wang2019}, we fine-tune paragraph BERT using the popular softmax-based listwise ranking loss. Listwise loss functions are generally more effective at ranking tasks than pointwise or pairwise loss functions \cite{Guo2019} and have been used successfully with neural ranking architectures \cite{Huang2013,Shen2014}.

Thus, a single training example for paragraph BERT consists of $n+1$ (quote query, source paragraph) input pairs -- one for the positive-labeled paragraph that contains the real quote the author chose and $n$ negative, unquoted, paragraphs from the same source document. Each of the $n+1$ packed input sequences is fed to BERT \textit{independently}. We use the final hidden vector C $\in R^h$ corresponding to the first input token $\mathtt{[CLS]}$ as the representation for each of the $n+1$ sequences, where $h$ is the size of the final hidden layer. We introduce one task-specific parameter, $V \in R ^ {h}$, whose dot product with $C$ is the score for each choice. We transform the $n + 1$ scores into probabilities using the softmax function: \\
\centerline{$P_i = \frac{e ^{V \cdot C_i}}{\sum_{p \in \mathbf{P}} e ^{V \cdot C_p}}$}\\ where \textbf{P} is the set of $n$ + 1 sequences. The loss is the negative log-likelihood (NLL) of the positive sequence. At inference time, for each quote query we create a (quote query, paragraph) packed input sequence for each paragraph in the corresponding single source document and rank them by their predicted scores.



In order to select the $n$ negative pairs for training, we sample paragraphs from the same source document as our positive, quoted paragraph. We explore three negative sampling methods -- uniform sampling, upweighting paragraphs with high TF-IDF cosine similarity to the quote query, and upweighting paragraphs positionally close to the positive paragraph in the source. 

\subsubsection{Span Prediction}

The span-level BERT model receives the same (quote query, paragraph) packed input sequence as the paragraph ranker. BERT processes and predicts spans in each paragraph independently. We use the final hidden vector $T_i \in R^h$ as the representation for each WordPiece in a given paragraph. We follow the standard approach of casting span prediction as two classification tasks, separately predicting the start and end of the span \cite{Devlin2019,Wang2019,Chen2017}.

We explore two different approaches for training span-level BERT -- \textbf{positive-only} \cite{Chen2017} and \textbf{shared-normalization} \cite{Clark2018a,Liu2019,Wang2019}. In the positive-only approach, span-level BERT is trained with only (quote query, paragraph) pairs that contain the quote the author selected. We discard all negative pairs from the training set, and span-level BERT is only trained on one pair at a time. We introduce start and end vectors, $S, E \in R^h$. The probability of word $i$ being the start of the quoted span is the dot product $S \cdot T_i$  followed by a softmax over all WordPieces in the example. We follow the same approach for calculating the probability of word $i$ being the end of the span using $E$. The loss is calculated as the average NLL of the correct start and end positions, i.e. the tokens in the paragraph the author actually quoted. Following \citeauthor{Devlin2019} \shortcite{Devlin2019}, at prediction time, the score of the span from position $i$ to $j$ is $S \cdot T_i + E \cdot T_j$. We consider a span valid if $j > i$ and $i$ and $j$ occur in the paragraph portion of the input. We retain a mapping from WordPieces to original tokens for prediction. Just as \citeauthor{Chen2017} \shortcite{Chen2017}, we adapt the positive-only model to the multi-paragraph setting at inference by running it on each paragraph in the corresponding source document independently, finding the maximum scoring span in each paragraph, then using the unnormalized score (before the softmax) to rank the spans. 

As first noted by \citeauthor{Clark2018a} \shortcite{Clark2018a}, applying this approach to the multi-paragraph setting may be problematic. As the model is trained on and predicts spans in paragraphs independently, pre-softmax scores across paragraphs are not guaranteed to be comparable. Further, since the model is only trained on positive paragraphs that contain true quote spans, it may not effectively learn to score spans in negative paragraphs. To resolve this, we utilize the most effective approach found by \citeauthor{Clark2018a} \shortcite{Clark2018a}, \textbf{shared-normalization}, which has also achieved SoTA performance with BERT-based QA systems \cite{Liu2019,Wang2019}. Just as for the paragraph model, we create $n + 1$ input sequences (1 positive, $n$ negatives), and each input is processed independently by span-level BERT. We then use a modified objective function where the softmax operation is shared across all positions in the $n+1$ pairs to calculate the start and end probabilities. Thus the probability of word $j$ in paragraph $i$ starting the span is: 

\centerline{$P_{ij} = \frac{e^{S \cdot T_{ij}}}{\sum_{p \in \mathbf{P}}\sum_{w \in p} e ^{S \cdot T_{pw}}}$}

\noindent
where $w$ iterates over each word in each of the $n$ + 1 examples. This forces the model to output comparable scores across multiple paragraphs and trains it to learn patterns effective in scoring both positive and negative paragraphs. The loss is the average NLL of the correct start and end positions in the positive paragraph. At prediction, the model is run on each paragraph in the corresponding source document independently, and the unnormalized scores are used to rank the top span in each paragraph. 

\begin{figure}
	\centering
	\includegraphics[width=0.47\textwidth]{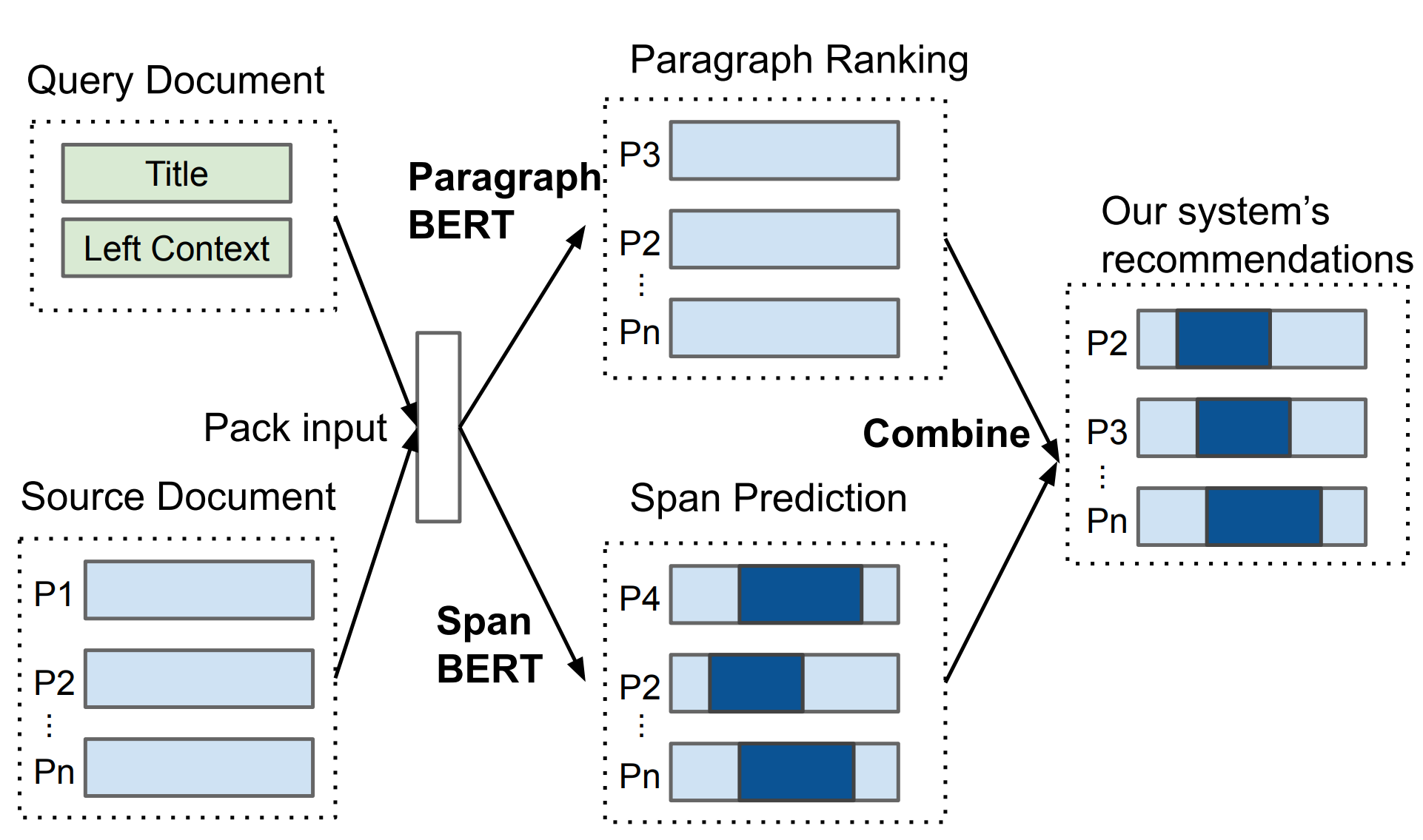}
	\caption{Combined BERT model architecture.}
	\label{fig:system}
\end{figure}

\subsubsection{Combined QR Model}

Due to the success in combining the scores of paragraph and span models in open-QA \cite{Wang2018a,Lee2018,Wang2019,Htut2018}, we explore a similar combination of the paragraph and span-level BERTs for our task. We hypothesize that the paragraph and span models learn to identify complementary aspects of quotable passages, so combining the scores will help rank paragraphs such as those which, although topically 
similar to the left context, contain differing amounts of quotable language, or paragraphs that contain very quotable phrases but are unrelated to the left context. Thus we create a single, aggregate score for a paragraph and its top ranked span at inference by combining the paragraph and span model scores using late fusion (Figure~\ref{fig:system}). We calculate the probability $p(p_i|q_j)$ of paragraph $p_i$ given quote query $q_j$ as the softmax over the scores of all possible source paragraphs for quote query $j$. The probability $p(s_{ik}|p_i,q_j)$ of span $s_{ik}$, the maximum scoring span in paragraph $i$, given quote query $q_j$ and its paragraph $p_i$ is calculated as the softmax over the maximum scoring span in each of the possible source paragraphs for $q_j$. Following \citeauthor{Lee2018} \shortcite{Lee2018}, we calculate the overall confidence of a paragraph $p_i$ and its predicted answer span $s_{ik}$ given the quote query $q_j$ as 

\centerline{score($s_{ik},p_i|q_j) = p(s_{ik}|p_i,q_j)^\alpha p(p_i|q_j)^\beta$} 

\noindent The importance of each score is determined by the hyperparameters $\alpha, \beta \in \{0, 0.5, ..., 10\}$, which we tune on our dev set. 

Alternatively, we could train a single model to jointly score paragraphs and spans. We tried training separate task-specific parameters for paragraph ranking and span prediction on top of a shared BERT model, then averaging the paragraph and span losses. However, this joint model has worse performance on our dev set, so we focus only on the combined, separately-trained model for our experiments. 

\section{Dataset} 
\label{sec:dataset}

\begin{figure*}
	
	\begin{subfigure}{.249\textwidth}
		\centering
		\includegraphics[width=\linewidth]{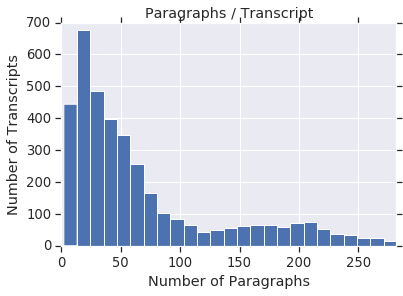}
		\caption{Paragraphs / Transcript}
		\label{fig:paragraphs_per_speech}
	\end{subfigure}%
	\begin{subfigure}{.249\textwidth}
		\centering
		\includegraphics[width=\linewidth]{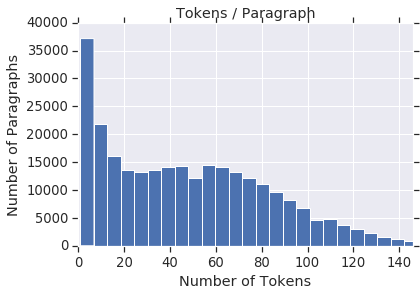}
		\caption{Tokens / Transcript Paragraph}
		\label{fig:tokens_per_paragraph}
	\end{subfigure}%
	\begin{subfigure}{.249\textwidth}
		\centering
		\includegraphics[width=\linewidth]{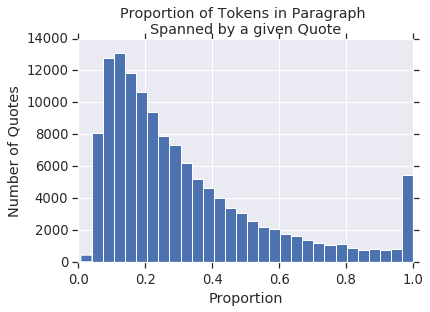}
		\caption{Fraction of paragraph quoted}
		\label{fig:quote_proportions}
	\end{subfigure}%
	\begin{subfigure}{.249\textwidth}
		\centering
		\includegraphics[width=\linewidth]{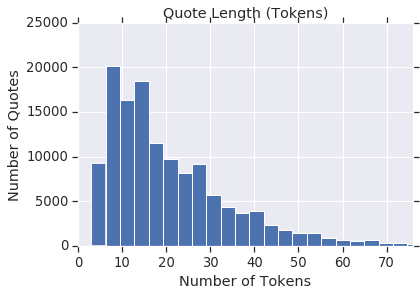}
		\caption{Quote Length (tokens)}
		\label{fig:tokens_per_quote}
	\end{subfigure}%
	\caption{Distributions over transcript, paragraph, and quote lengths.}
\end{figure*}

We use the QUOTUS dataset for our experiments \cite{Niculae2015}. The dataset consists of two sets of texts -- transcripts of US Presidential speeches and press conferences from 2009-2014 and news articles that report on the speeches. The authors identify the quotes in each news article that originate from the transcripts. They release the transcripts and the collection of aligned quotes, containing the text of the quote in the news article, its aligned position within the source transcript, and the corresponding news article metadata (title, url, timestamp).\footnote{\url{http://snap.stanford.edu/quotus/#data}}

We crawled the provided news article URLs on Mar 15, 2019 and fwere able to download 110,023 news articles ($\approx 75\%$ of the provided URLs). We extracted the title and body content of each article using BeautifulSoup\footnote{\url{https://www.crummy.com/software/BeautifulSoup/bs4/doc/}} and
removed duplicates, articles with less than 100 characters of body content, and articles where we could not locate the quotes.
This leaves our final dataset of \textbf{131,507 quotes from 3,806 different White House transcripts across 66,364 news articles}. The quotes are grouped into 58,040 clusters, groups of quotes, possibly with minor edits, which all align to roughly the same place in the transcripts. We then preprocessed articles and transcripts with CoreNLP's sentence segmenter and word tokenizer \cite{Manning2014}. 

We use the alignments between the transcripts and news articles to create the labels for our dataset, learning to recommend paragraphs and spans from a source transcript to a journalist as they write a news article. For each aligned quote, we construct the quote query as described in \S\ref{sec:quote_rec} (news article title + left context leading up to the specific quote). 
The \textbf{positive span} is the exact span of tokens in the transcript that are aligned to the quote in the article. The \textbf{positive paragraph}(s) are those transcript paragraph(s) that contain the positive span ($\approx 99.5\%$ of positive spans are contained in one single paragraph and do not cross paragraph boundaries).  The average transcript contains 71 paragraphs, (median 43, Figure~\ref{fig:paragraphs_per_speech}),
and the average paragraph is 49 tokens long (median 45, Figure~\ref{fig:tokens_per_paragraph}). Similar to findings in studies of media coverage of political debates \cite{Tan2018}, paragraphs earlier in the speech tended to be quoted more often than those later.
A positive span, on average, covers 32\% of the corresponding positive paragraph (median 24\%, Figure~\ref{fig:quote_proportions}) and is 21 tokens long (median 17, Figure~\ref{fig:tokens_per_quote}). Only 3\% of quotes span an entire paragraph. Many positive spans either start at the beginning (13\%) or end at end (32\%) of the positive paragraph. Otherwise, span start positions are relatively uniformly distributed over the first 80\% of a given paragraph and end positions over the last 80\%.


\begin{table}
	\caption{Preprocessed dataset summary statistics.}
	\label{table:split_sizes}
	\small
	\centering
	\begin{tabular}{r|rrr}
		
		& \# Quote & \# Speech &  \# Article \\ \hline
		Train & 94,931    & 2,823     & 47,698      \\ 
		Dev   & 15,472    & 391      & 7,763     \\ 
		Test  & 21,104    & 592     & 10,903    \\ 
	\end{tabular}
\end{table}

We split our dataset into train, dev and test by date, splitting by the transcript publication date. We select transcripts from 2009-01-01 to 2012-11-07 as our training set, 2012-11-08 to 2013-06-19 as our dev set, and 2013-06-20 to 2014-12-31 as our test set (see Table \ref{table:split_sizes}). 
The split dates were selected to ensure no news articles quote from transcripts in different data splits. 

Nearly all ($93\%$) of the news articles in our dataset contain only quotes which all align to the same source transcript. This is reasonable as a journalist is much more likely to quote from the most recent and topically relevant source speech than search over less relevant and up-to-date speeches \cite{Niculae2015}. 

For the $0.5\%$ of quotes in the training set that span multiple paragraphs $(\approx 500$ quotes), we create separate span and paragraph training examples for each part, independently sampling negative paragraphs for each. For evaluation, we positively label all $p$ spanned paragraphs, concatenate the spans across the $p$ positive paragraphs as the positive span, and concatenate the predicted spans from the top $p$ ranked paragraphs to create the predicted span.

\section{Model Experimental Settings}
\label{sec:experimental_settings}

\subsubsection{Parameter Settings} We cap titles, left contexts and source document paragraphs  at 20, 100 and 200 WordPieces, respectively. In \S\ref{ablation}, we explore the effects of decreased article context on predictive accuracy. We fine-tune the publicly released 
BERT Base\footnote{\url{https://storage.googleapis.com/bert_models/2018_10_18/uncased_L-12_H-768_A-12.zip}}
for our paragraph and span models and perform a search over batch size $\in \{16, 32\}$ and learning rate (Adam) $\in \{5e-5, 3e-5, 2e-5\}$. We train our models for a maximum of 4 epochs and select hyperparameters and perform early stopping using our dev set. We also perform a search over $n \in \{3, 6, 9, 12\}$ sampled negative paragraphs per positive paragraph for our paragraph and shared-norm models. In preliminary experiments on our dev set, we find no significant differences between uniform random sampling and our proposed non-uniform sampling schemes. Instead, model performance generally increases by sampling a larger number of negatives. We suspect this is because a larger sample will likely include both hard- and easy-to-rank negatives, thus training the model to differentiate between the positive paragraph and both semantically similar and dissimilar paragraphs. The best paragraph model is trained with 12 negative examples, batch size 32, and learning rate 2e-5. The best shared-norm span model is trained with 9 negative examples, batch size 16, and learning rate 3e-5. The best positive-only span model is trained with batch size 32 and learning rate 3e-5.

\subsubsection{Metrics} To evaluate paragraph rankings, we use mean average precision (mAP) and top-$k$ accuracy, defined as the proportion of examples where the positive paragraph is ranked in the top $k$. Since only $\approx 100$ quotes in the test set span multiple positive paragraphs, mAP is nearly equivalent to mean reciprocal rank. For top-k accuracy, we require that at least 1 of the positive paragraphs is ranked in the top $k$. 

Since the setup for our span prediction task is identical to QA, we evaluate the span-level models using the same two QA metrics that we used to evaluate the expert predictions in the pilot study (\S\ref{sec:pilot}) -- exact match (EM), and macro-averaged bag-of-words F1. EM measures if the predicted span exactly matches the positive quote, and BOW-F1 measures their average word overlap. 
We evaluate the span models in two settings, \textit{positive} and \textit{top}. In the \textit{positive} setting, we assume that an author has already selected the positive paragraph and is interested in the recommended span for that paragraph. Thus, for a given test example, we only evaluate our model's predictions on the positive, quoted paragraph(s) against the true quote. This is equivalent to a closed, single paragraph QA evaluation. In the \textit{top} setting, for a given quote, we run the model on all possible paragraphs in the corresponding source, identify the top scoring span in each paragraph, rank the spans by their scores, and evaluate the top ranked span. This setting is equivalent to open-QA evaluation. Though more realistic, this is harder, as the model may not rank the predicted span in the positive paragraph the highest. 

Since, as noted in the pilot study (\S\ref{sec:pilot}), there are often multiple relevant paragraphs and quotes for a given context, evaluating models in these single-answer settings is overly strict. Thus, in \S\ref{sec:crowdsource}, we perform a large-scale crowdsource evaluation of our best model's predictions, assessing whether its ``incorrect'' recommendations are still relevant and appropriate.




\subsubsection{Paragraph Ranking Baselines} We compare our paragraph ranker against 2 common bag-of-words retrieval models -- \textbf {TF-IDF} cosine similarity and \textbf{BM25} \cite{Jones2000}. We also evaluate performance of \textbf{Doc2Vec} cosine similarity \cite{Le2014} to see if it can capture semantic similarity between the left context and paragraph. Finally, we evaluate two neural ranking baselines -- \textbf{K-NRM} \cite{Xiong17}, a popular kernel pooling neural network that uses a translation matrix of word embedding similarities between a query and document to learn-to-rank documents for ad-hoc retrieval, and \textbf{Conv-KNRM} \cite{Dai2018}, which builds on K-NRM by using CNNs to model soft n-gram matches between a query and document. 

As noted in \S\ref{sec:related_work}, previous approaches to QR use different tasks setup (e.g. ranking a predefined list of quotes), and the proposed models use additional, non-textual features, such as popularity data crawled from the web \cite{Tan2015,Tan2016}. Similarly, the SoTA citation recommendation system \cite{Jeong2019} makes use of citation network metadata features in addition to BERT. Due to these incompatibilities with our QR task, where models are only provided with the texts of the context and source, we do not evaluate these previous systems.

For each baseline, we calculate the similarity between a quote query and each possible paragraph in the corresponding source document, just as for our BERT models. For Doc2Vec we use the distributed bag-of-words algorithm and search over the hyperparameters suggested by \citeauthor{Lau2016} \shortcite{Lau2016}. For K-NRM and Conv-KNRM, we use the suggested hyperparameters from \citeauthor{Xiong17} \shortcite{Xiong17} and \citeauthor{Dai2018} \shortcite{Dai2018} and use pre-trained Glove embeddings \cite{Pennington2014}. We train the models on our training set and use early stopping on our dev set. For each baseline we search over the number of tokens of left context $\in \{10, 20, 30, ..., 100\}$ to include in the article query in addition to the title, and find $30$ tokens works best for Conv-KNRM and $40$ for all other baselines on the dev set. 

\section{Results \& Analysis}
\label{sec:results}

\subsection{Paragraph Ranking}

\begin{table}
	\caption{Paragraph ranking results. \textbf{Bold} entries significantly outperform all other models (p $<$ 0.01). Models 6-10 are BERT models. Acc is short for top-k accuracy.}
	\label{table:ranking_results}
	\small
	\begin{tabular}{l|llll}
		
		Method & mAP            & Acc@1            & Acc@3            & Acc@5            \\ \hline
		1. TF-IDF  &  39.0    & 27.3   &   42.5       &   50.7       \\ 
		2. BM25  &  39.9      &  28.4    & 43.5      &    51.5       \\ 
		3. Doc2Vec & 31.7    & 21.1      & 33.4          & 41.4               \\
		4. K-NRM & 39.2       &   27.3        &   42.6          & 51.1              \\
		5. Conv-KNRM &    38.9    &  27.2         &    41.9         &   50.4            \\ \thickhline 
		6. Paragraph   &  52.4       &   39.7       &    58.1      &    67.3      \\ \hline
		7. Positive-only Span     & 32.2          & 20.0           & 34.3          & 44.2          \\ 
		8. Shared-norm Span     & 51.3          & 38.6          & 56.8          & 65.9          \\ \hline 
		9. Combined (6 + 7)   &  52.7       &  40.0        &  58.1        &    67.7      \\ 
		10. Combined (6 + 8) & \textbf{53.2} & \textbf{40.5} & \textbf{59.1} & \textbf{68.2}
		
	\end{tabular}
\end{table}

Table~\ref{table:ranking_results} shows the paragraph ranking results for all the methods with their best settings. We test significance with a permutation test, p $<$ 0.01. We find that paragraph BERT (\#6) significantly outperforms all baselines (\#1-5) by a large margin, outperforming BM25, the best baseline by $\approx 13\%$ absolute and $30-40\%$ relative across all metrics. Examining the baselines, we find that Doc2Vec is much worse than the other baselines that can model exact token matches between passages. We hypothesize that these exact matches, particularly of named entities, are a strong indicator of transcript paragraph relevance. For instance, if the journalist is discussing health care and mentions the Affordable Care Act in the left context, then paragraphs that not only are topically related to health care, but also contain that specific entity may be more relevant. BERT's ability to model both exact and semantically similar matches between a context and paragraph across multiple levels of granularity might lead to its improved performance over the token and embedding matching baselines. 

However, lexical and semantic similarity to the context is only one aspect of a paragraph's quotability \cite{Mizil2012}. In order to be selected by a journalist, a paragraph must not only be topically relevant, but also contain material that is interesting, meaningful, and serves the journalist's purpose. All of our baselines rank paragraphs solely by similarity to the context and fail to model the quotability of the paragraph. BERT's superior performance in scoring hard-to-rank paragraphs, such as those which all discuss the same event, person, or topic as the left context, but contain varying amounts of quotable language, indicates that it may also capture aspects of quotability beyond relevance. 


As the span-level BERT models can also be used to rank paragraphs, we evaluate the positive-only (\#7) and shared-norm (\#8) span models on the paragraph ranking task. For a given quote, we run the span models on all possible paragraphs from the corresponding source and rank each paragraph 
by its maximum scoring span.
Though training with shared-norm loss substantially improves ranking performance over the positive-only model, paragraph BERT (\#6) significantly outperforms both span BERT models across all metrics. However, although paragraph BERT is overall a stronger ranker, it does not outperform shared-norm BERT on every test example. On 175 examples, shared-norm BERT ranks the ground truth as the top 1, but paragraph BERT ranks it outside the top 5. We hypothesize that this is due to differences in the granularities of their objectives, with each model 
identifying different aspects of what makes a paragraph quotable and relevant. 

Finally, our experiments combining the paragraph and span models parallel findings in open-QA \cite{Wang2019}, with the combined paragraph and shared-norm model yielding the best performance (\#10) and significantly outperforming all other models across all ranking metrics. It improves over the single paragraph BERT model by $\approx 1\%$ absolute across each metric and by $\approx 15\%$ over BM25, the best baseline. This combined model upweights the paragraph score 
relative to the shared-norm score: $\alpha = 3, \beta = 9.5$. To evaluate if ranking performance can be boosted even while adding a weaker span ranker, we also experiment with combining the paragraph model and positive-only span model. The best performing combined positive-only model (\#9) upweights the paragraph score $\alpha = 1.5, \beta = 7.0$. Though it does not perform as well as the combined shared-norm model, the combined positive-only model improves over the single
paragraph model in mAP and top 1 and 5 accuracy. This further validates our hypothesis that the span and paragraph-level models learn complimentary aspects of quotable paragraphs.

\subsubsection{Error Analysis} 
\begin{figure}
	\centering
	\includegraphics[width=.75\linewidth]{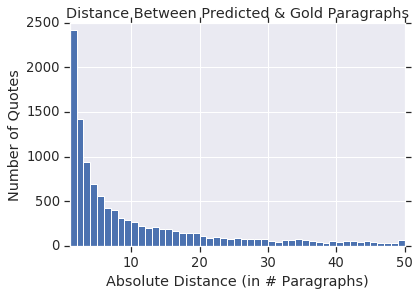}
	\caption{Distribution over absolute distance (in number of paragraphs) between the top paragraph incorrectly recommended by the combined BERT model and the positive paragraph chosen by the original journalist.}
	\label{fig:differences}
\end{figure}

To better understand the errors our best combined BERT model (\#10) makes, we examine the 12,557 quotes in the test set where it does not rank the positive paragraph as the top 1. We focus our analysis on the top-1 predicted (incorrect) paragraph, and measure the absolute distance (in \# of paragraphs) between it and the positive, actually quoted paragraph. As can be seen in figure \ref{fig:differences}, we find that our model often recommends a paragraph nearby the one selected by the journalist. For approximately $20\%$ of quotes, the model recommends a paragraph directly adjacent to the positive paragraph, and for roughly $38\%$, it recommends a paragraph within 3 positions of the positive. As adjacent paragraphs in a speech are more likely to be on the same topic, this confirms that our model's incorrect recommendations are often still topically similar to the paragraph the journalist chose.   

\begin{table}
	\caption{Span prediction results of BERT models. \textbf{Bold} entries significantly outperform all other models.}
	\label{table:span_results}
	\centering
	\small
	\begin{tabular}{l|llll}
		Method & \multicolumn{1}{c}{\begin{tabular}[c]{@{}c@{}}EM\\ \textit{Positive}\end{tabular}} & \multicolumn{1}{c}{\begin{tabular}[c]{@{}c@{}}EM\\ \textit{Top}\end{tabular}} & \multicolumn{1}{c}{\begin{tabular}[c]{@{}c@{}}F1 \\ \textit{Positive}\end{tabular}} & \multicolumn{1}{c}{\begin{tabular}[c]{@{}c@{}}F1 \\ \textit{Top}\end{tabular}} \\ \hline
		
		1. Paragraph (Full) & 2.9 & 0.6 & 42.2 & 24.4  \\ 
		2. Paragraph (1st-Sent) & 4.8 & 1.2 & 28.3 & 10.7 \\
		3. Paragraph (Last-Sent) & 5.5 & 1.4 & 30.6 & 12.8 \\ \hline
		
		4. Positive-only Span         & 21.0 & 7.8  & 57.4  & 23.6  \\ 
		
		5. Shared-norm Span 
		& \textbf{22.8}  & 12.0  & \textbf{59.0} & 34.1  \\ \hline
		
		6. Combined (1 + 4) & 21.0 & 11.2 & 57.4 & 33.8 \\ 
		7. Combined (1 + 5)
		& \textbf{22.8}  & 12.0  & \textbf{59.0} & \textbf{34.5}  
	\end{tabular}
\end{table}

\subsection{Span Prediction}

As seen in Table \ref{table:span_results}, span prediction is harder than paragraph ranking. As baselines, we use the entire paragraph (\#1), its first sentence (\#2), or its last sentence (\#3) as the predicted span. These paragraphs are either the positive paragraph (positive setting) or the top ranked paragraph (top setting) by the best single paragraph model (ref. Table~\ref{table:ranking_results} \#6). The positive-only span model (\#4) significantly outperforms the baselines in EM \textit{positive} \& \textit{top} and F1 \textit{positive}, but its predictions are worse than the full paragraph on F1 \textit{top}. Parallel to findings in open-QA \cite{Clark2018a,Wang2019}, the shared-norm model (\#5) significantly outperforms the positive-only model and baselines across all metrics, even EM and F1 \textit{positive}, demonstrating that the shared loss function forces the span model to produce more comparable scores between different paragraphs and improves its ability to identify spans in positive paragraphs. 

Just as with findings in open-QA and our paragraph ranking task, combining the paragraph and shared-norm models improves performance (\#7). Note, the performance in the \textit{positive} setting is identical as we only compute overall confidence scores for the \textit{top ranked} span in each paragraph. The paragraph score will not influence the selection of a span in a given paragraph, only the ranking of those spans across paragraphs. The differences between the best combined shared-norm model (\#7) and the single shared-norm model (\#5) are not as pronounced as in the ranking task. The combined shared-norm model only significantly outperforms the single shared-norm model on F1 \textit{top} by $1\%$ relative, with no improvement on EM \textit{top}. We also evaluate the combined positive-only model (\#6). Though worse than both shared-norm models, it improves $\approx 40\%$ relative over the single positive-only model in EM and F1 \textit{top}.



\subsubsection{Error Analysis}
\label{sec:span_error_analysis}

To further assess the combined shared-norm model (\#7), we examine its errors on the test set, comparing its predictions on the positive paragraph to the true quote. In some instances, we find that the model recommends a span that is semantically similar to the true quote or one that is quoted multiple sentences later. For 15\% of errors, we find that the model under-predicts, predicting a sub-phrase in the true quote. However, parallel to the pilot study findings (\S\ref{sec:pilot}), the most common error is over-prediction (47\% of errors), where the model predicts a span containing the true quote and some phrases surrounding it. 

Further, as in \S\ref{sec:pilot}, we find that in many instances of over-prediction the author paraphrases the predicted surrounding phrases in the introduction to or discussion of the quote. For instance, on the example shown in Table~\ref{table:example_datapoint}, the model over-predicts ``At one point, they won 27 games straight- the second-longest winning streak ever, \textit{extraordinarily impressive- almost as impressive as the Bulls' 72-win season}'' (true quote \textit{italiced}). Though incorrect, the non-quoted phrase is still relevant since the journalist describes the team's successes in the context introducing the quote. 


\subsection{Crowdsourced Evaluation}
\label{sec:crowdsource}

As noted in the findings of our pilot study and error analysis, there are often multiple quotes and paragraphs a journalist could select for a given context. Thus we hypothesize that our model's ``incorrect'' top recommendations (non-positive quotes) could still be relevant and appropriate. Due to the limited time of the two journalists, we turn to crowdworkers to perform a large-scale evaluation our model's predictions. 

We focus this evaluation on paragraph rather than span predictions since we believe authors writing a new document will use the entire paragraph to 1) better understand the broader source context and 2) once a specific quote is chosen, paraphrase other content from the paragraph to introduce and contextualize it. 
We randomly sample 1,000 quotes from our test set where our best combined model does not rank the positive paragraph as the top 1 and evaluate, using Mechanical Turk crowdworkers, whether the model's top ranked (non-positive) paragraph is relevant and quotable.   

For each quote, crowdworkers see the article title, left context (100 WordPieces, converted to regular tokens and rounded up to a full sentence), and the top non-positive paragraph ranked by our best model. They are asked ``\textit{Does this paragraph contain an excerpt that would be appropriate to quote in the next sentence of the news article?}'' 
We block workers who do not follow instructions and collect 5 judgments per question. Each worker labeled 10 questions, for a total of 500 unique workers. Inter-annotator agreement is 0.26 (Fleiss' $\kappa$), indicating fair agreement. We aggregate labels using majority vote. 

Workers mark our predicted paragraph as relevant 85\% of the time, suggesting that our model's top predictions seem quotable and meaningful to non-experts even if they are not the paragraph chosen by the original journalist. 
Table~\ref{table:example_datapoint_crowdsource} shows an example context and top ranked paragraph, along with the true next sentence in the original article. The paragraph is marked as quotable by all 5 workers, and, in fact, the selected quote in the original article is similar to the last sentence of the top ranked paragraph.

\begin{table}[t]
	\caption{A top ranked (non-positive) paragraph that is marked as quotable by participants in our crowdsourced evaluation. The paragraph is topically relevant to the document context and contains many specific figures and anecdotes. Its \uwave{last sentence}, just as the ground truth quote, discusses Yellen's relationship to everyday Americans.}
	\small
	\begin{tabular}{|p{8cm}|}
		\hline
		\textbf{Document Title:} Janet Yellen to become the first female US Fed boss        \\ 
		\textbf{Document Context:} ... Yellen will chair the Federal Reserve board, replacing Ben Bernanke after his second four-year term comes to an end at the beginning of next year.
		\\ 
		\textbf{Top Ranked Source Paragraph:} So, Janet, I thank you for taking on this new assignment. And given the urgent economic challenges facing our nation, I urge the Senate to confirm Janet without delay. I am absolutely confident that she will be an exceptional chair of the Federal Reserve. I should add that she'll be the first woman to lead the Fed in its 100-year history. \uwave{And I know a lot of Americans- men and women- thank you for not only your example and your excellence, but also being a role model for a lot of folks out there}.	 \\
		\textbf{True Next Sentence:} ``American workers and families will have a champion in Janet Yellen,'' President Obama said.
		\\ \hline
	\end{tabular}
	\label{table:example_datapoint_crowdsource}
\end{table}

\subsection{Ablation Studies}
\label{ablation}

To explore the effects of decreasing the amount of article content, we perform 2 ablation experiments, training new paragraph and span models and combining their scores using the same hyperparameters as our best models. For brevity, Table \ref{table:ablation} shows results for only a subset of the ranking and span prediction metrics. Decreases across all metrics are consistent with the shown subset. First, we remove the article title, but keep 100 WordPieces of left context. Performance (\#2) decreases by $\approx 2\%$ and $1\%$ absolute across ranking and span metrics, respectively. This decrease is expected as an article title provides both a high level overview of the article and the only context when the quote occurs in the beginning of the article. However, in domains such as news, where the title is often written after the body of the article, performance on this experiment may be more realistic. Next, we keep 20 WordPieces of title content, but vary the left context WordPieces $\in \{50, 25, 10\}$. Performance across all metrics decreases with context size, with the sharpest drops from 25 to 10 WordPieces. Encouragingly, for data and tasks where the left context size is limited, with only 50\% as much left context as the full model, the 50 WordPiece model (\#3) only loses $\approx$1-2\% relative performance across each metric. Though these models can use fewer tokens, the tokens are at the end of the context right before the quote. We hypothesize that, due to this proximity, they are likely to contain strong signals for prediction.


\begin{table}
	\caption{Performance of combined model on dev set with varying amounts of quote query context. The full model is our best performing combined model that uses the title and 100 WordPiece tokens as the left context. Acc is short for top-k accuracy. \textit{Pos} is short for \textit{positive.} ``-'' indicates performance relative to our best model.}
	\label{table:ablation}
	\small
	\centering
	\begin{tabular}{l|cc|cc}
		& \multicolumn{2}{c|}{Paragraph Ranking} & \multicolumn{2}{c}{Span Prediction} \\ 
		& Acc@1 & Acc@5 & F1 (\textit{Pos}) & F1 (\textit{Top}) \\ \hline
		
		1. Full Model & 40.6 & 68.8 & 59.0 & 34.5 \\ \thickhline 
		2. No Title & - 1.9 & - 2.0 & - 0.7 & - 1.2  \\ \hline
		3. Context: 50 & - 1.0 & - 1.3 & - 0.4 & - 0.5 \\ 
		4. Context: 25  & - 2.7 & - 3.5 & - 1.1 & - 1.7  \\ 
		5. Context: 10 & - 8.3  & - 9.0  & - 2.7  & - 5.3 \\
	\end{tabular}
\end{table}

\section{Conclusion and Future Work}
\label{sec:discussion_future_work}

In this paper, we introduce a new context-based quote recommendation task, using the content an author has already written to rank the paragraphs and score the spans of a source document by their predicted relevance and quotability. We cast the task as a variant of open-domain question answering, and explore applications of state-of-the-art BERT models to the problem.
Though we conduct experiments on a dataset of speeches and news articles, our proposed approach could be used in many domains. Bloggers, students, and journalists could use our method to rank passages in source documents of any kind, such as press releases, meeting transcripts, and book chapters.

For domains where authors may quote from multiple source documents with a very large number of total paragraphs, we could explore how the BERT model scales and determine whether we would need to train a document ranker to filter documents as a first step. 
For applications where the right context may be partially written before selecting a quote, we could explore the effects of including bidirectional context on predictive performance. 
Finally, to improve identification of quotable content, we could experiment with adding manually designed quotability features to BERT's learned representations (e.g. use of positive or negative words or personal pronouns) 
\cite{Mizil2012,Bendersky2012,Tan2018} or learn them automatically by training a language model on a large 
collection of quotes, such as Memetracker \cite{Leskovec2009}.

\section*{Acknowledgements}
We would like to thank journalists Megan Chan and Ashley Edwards for their participation in our pilot study and helpful feedback. We also thank Jason Baldridge and the anonymous reviewers for their insightful comments that improved the paper.

\fontsize{9.0pt}{10.0pt} 
\selectfont
\bibliographystyle{aaai}

\end{document}